\documentclass{article}

     \usepackage[final]{neurips_2020}

\usepackage[utf8]{inputenc} % allow utf-8 input
\usepackage[T1]{fontenc}    % use 8-bit T1 fonts
\usepackage{hyperref}       % hyperlinks
\usepackage{url}            % simple URL typesetting
\usepackage{booktabs}       % professional-quality tables
\usepackage{amsfonts}       % blackboard math symbols
\usepackage{nicefrac}       % compact symbols for 1/2, etc.
\usepackage{microtype}      % microtypography
\usepackage{pifont}
\newcommand{\xmark}{\ding{55}}%

\usepackage{xurl}
\usepackage{graphicx}
\usepackage{tabularx}
\usepackage{enumitem}

\title{Assessing out-of-domain generalization for robust building damage detection}

\author{%
  Vitus Benson \\
  Institute of Computer Science \\
  Campus Institute Data Science \\
  University of Goettingen \\
  \texttt{vitus.benson@icloud.com}  \\
  \And Alexander Ecker\\
  Institute of Computer Science \\
  Campus Institute Data Science \\
  University of Goettingen \\
  \texttt{ecker@cs.uni-goettingen.de}  \\
}

\begin{document}

\maketitle

\begin{abstract}
% Disaster management aims at limiting the negative impact of natural disasters. 
An important step for limiting the negative impact of natural disasters is rapid damage assessment after a disaster occurred. For instance, building damage detection can be automated by applying computer vision techniques to satellite imagery. Such models operate in a multi-domain setting: every disaster is inherently different (new geolocation, unique circumstances), and models must be robust to a shift in distribution between disaster imagery available for training and the images of the new event. Accordingly, estimating real-world performance requires an out-of-domain (OOD) test set. However, building damage detection models have so far been evaluated mostly in the simpler yet unrealistic in-distribution (IID) test setting. Here we argue that future work should focus on the OOD regime instead. We assess OOD performance of two competitive damage detection models and find that existing state-of-the-art models show a substantial generalization gap: their performance drops when evaluated OOD on new disasters not used during training. Moreover, IID performance is not predictive of OOD performance, rendering current benchmarks uninformative about real-world performance. Code and model weights are available at \url{https://github.com/ecker-lab/robust-bdd}.%For a simple ResNet-based model with a particularly large generalization gap, we show that Stochastic Weight Averaging and Adaptive Batch Normalization can decrease this gap. 
\end{abstract}

\section{Introduction}
% Intro to the subject
\paragraph{Building damage detection.} Climate change increases the frequency and severity of natural disasters \citep{ipcc2012}. Hence, improving the disaster management cycle will have a positive impact. One component of this cycle that recently has been in focus of researchers is automatic damage assessment. The International Charter Space and Major Disasters\footnote{\url{https://disasterscharter.org/}} regulates very high resolution satellite imagery to quickly be made available. Based on this data, mapping products up to building-level can be generated. Building damage detection (BDD) refers to the task of translating satellite imagery into maps with building damage levels. Deep neural networks (DNNs) have been applied to automate BDD, which makes resulting mapping products quicker available and standardizes the damage assessment.

% What has been done
\paragraph{Previous work.} BDD has been understood in three main ways: \emph{(1)}~as change detection \citep{doshi2018}, \emph{(2)}~as patch-wise image classification \citep[][among others]{fujita2017, xu2019} or \emph{(3)}~as semantic segmentation \citep[][among others]{bai2018, ghaffarian2019}. DNNs for semantic segmentation in the BDD context usually employ an encoder-decoder architecture, either in a one-stream (fusing input data along the channel dimension) or in a two-stream \citep[encoding pre- and post-disaster imagery separately;][]{daudt2018} configuration. Since time is valuable in BDD, researchers have leveraged data from different sensors, so that the first incoming data can be used for early estimates and later predictions use the maximum available information \citep[][among others]{duarte2018, adriano2019, rudner2019}.

% xBD
\paragraph{The xBD dataset.} Driven by the recent xView 2 challenge\footnote{\url{https://xview2.org/}}, the research focus has shifted more towards tackling BDD as semantic segmentation of very-high resolution (<\,5\,m/pixel) satellite imagery. Along with the challenge, the xBD dataset \citep{gupta2019} was released. It contains annotated satellite imagery at 0.5,m ground resolution from 19 natural disasters across the globe (Fig.~\ref{fig:map}). Each of the 22,068 samples is a combination of a pre- and a post-disaster RGB satellite image of size $1024 \times 1024$ pixels together with building polygons with associated damage levels. A unified damage grading scheme across disasters has been developed allowing for image pixels to be assigned one of five labels: background, undamaged, minor damage, major damage or destroyed. The damage grades refer only to building damage. Model performance is measured using the xView 2 score (a weighted combination of F1 scores~\cite{gupta2019}) on a so-called holdout set, which is an IID test set containing new samples from disaster instances that are part of the training set.

% Cite other xBD related publications here ????

\section{Measuring generalization}
\begin{figure}[t]
    \centering
    \includegraphics[width=\textwidth]{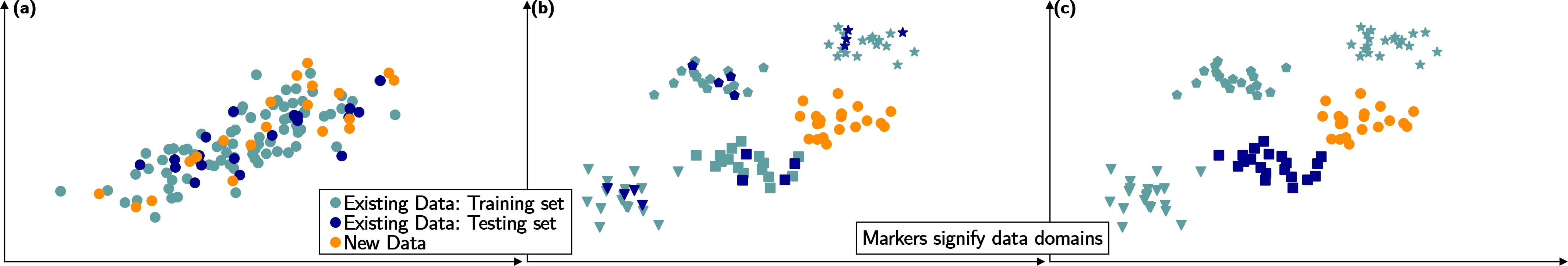}
    \caption{(a) Setting where IID testing estimates generalization accurately. (b) Setting where IID testing will overestimate generalization. (c) OOD testing will provide more accurate estimates of generalization for that setting.}
    \label{fig:iidood}
\end{figure}

% IID
% OOD maybe + viz
\paragraph{IID and OOD performance.}
For real-world applications, machine learning models need to generalize to new, previously unseen, data. This capability to generalize should be the ranking criterion during model selection. To estimate it, we have to make assumptions. If we assume new data samples to be independent and drawn from the same distribution as the existing data (IID), a legitimate criterion is measuring model performance by splitting the existing data randomly into a training and a test set and then estimating performance on the (IID) test set (Fig.~\ref{fig:iidood}a). 

However, in many real-world scenarios -- including BDD -- data samples are not entirely independent, and data used during training does not always come from the same distribution as that encountered later during deployment. For instance, in the xBD dataset, images are collected and annotated from a relatively small number of natural disasters and images from within one disaster are likely to be more similar than images across disasters (illustrated in Fig.~\ref{fig:iidood}b; each disaster is represented by a different symbol). In such a situation, splitting the existing dataset randomly (IID) into training (teal) and test (blue) set may not be predictive of the performance on new data (orange). Instead, one can likely obtain more robust estimates of OOD performance by splitting across groups of data samples (i.e. disasters; Fig.~\ref{fig:iidood}c).

% Multi-domain
% Previous BDD gen work
\paragraph{Previous work.} Using the OOD test setting for evaluating BDD is not yet common. \citet{xu2019} report a drop in performance when training on two earthquakes and testing on a third one from a different region. \citet{li2020} study the performance of domain adaptation methods in the context of transferring damage models trained on airborne imagery of one hurricane to another. \citet{gupta2020} report mediocre OOD generalization of their xView 2 challenge model. Concurrent to our work, \citet{valentijn2020} study the generalization performance of patch-wise BDD when target building polygons are provided. They report that training models only on disasters from one damage-driving force (wind or water) improves generalization to new disasters of the same type.

\section{Methods}
\begin{figure}
    \centering
    \includegraphics[width = \textwidth]{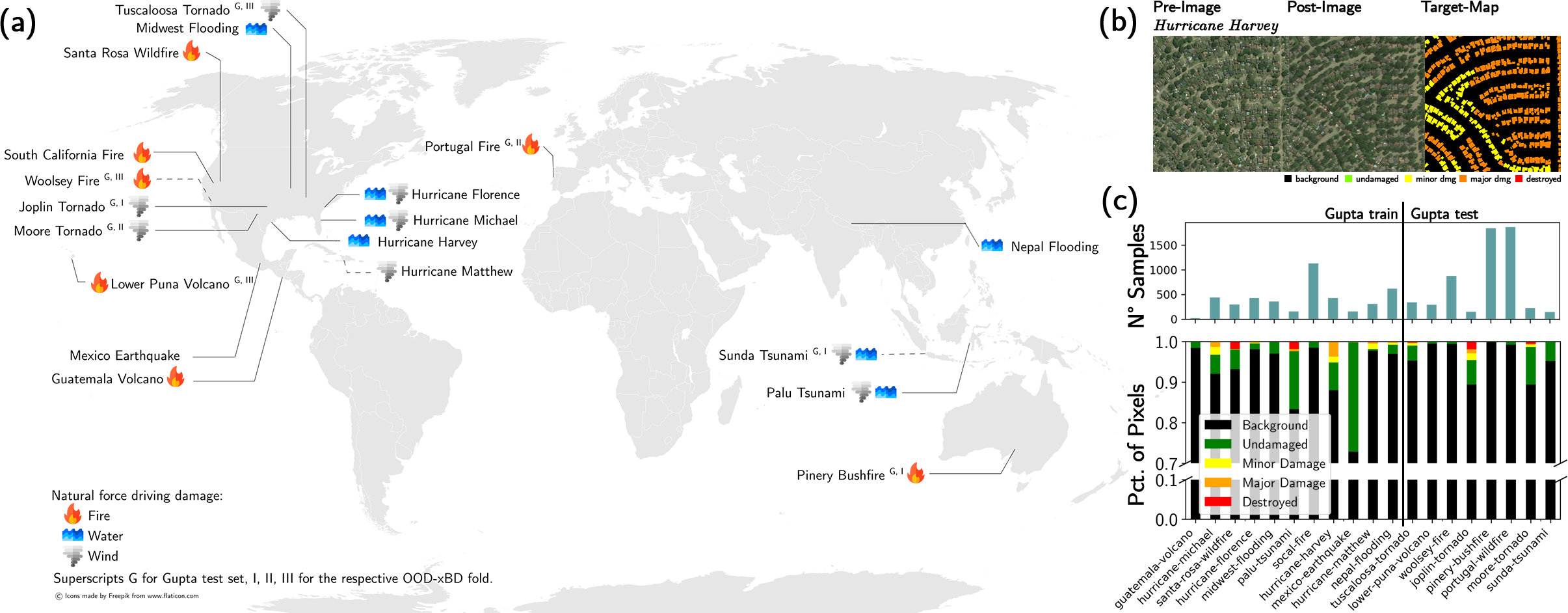}
    \caption{Panel (a) shows a map of disasters in xBD and their test set association, panel (b) shows a sample from xBD and panel (c) shows the number of samples and the pixel-distribution per disaster.}
    \label{fig:map}
\end{figure}

% Gupta split maybe + viz
% OOD-xBD split
\paragraph{Testing OOD on xBD.}
Fig.~\ref{fig:map} visualizes xBD and the contained disasters. \citet{gupta2020} create a test set containing the so-called tier 3 data -- eight natural disasters added to xBD at a later stage of the xView 2 challenge. We refer to this test set as the ``Gupta test set.'' It represents a challenging OOD test case, since the amount of training data compared to the full xBD dataset is relatively small. The advantage of this choice of disasters is that the training set contains the same eleven disasters as the original xBD holdout set, thus remaining compatible with using the original xBD holdout set as an IID test set.

The Gupta split tests true generalization. Yet, it is rather wasteful with training data as it holds back almost half of the dataset for testing. In production, models would likely use all available data for training and then predict just one new instance. The common strategy from statistics here is leave-one-out cross validation, which maximizes training data and provides uncertainty measures of the estimated performance. However, with 19 disasters, full leave-one-out cross validation would be computationally very demanding. We therefore define OOD-xBD as a test regime with three data folds. It allows for the use of more training data, is computationally more tractable, provides uncertainty estimates and can be used in combination with the xBD IID test set.
The disasters in the test sets of the three folds we chose are as follows:
\vspace{-\parskip}
{
\begin{itemize}[noitemsep,topsep=6pt,partopsep=0pt]
    \item Fold 1: Joplin tornado, Pinery bushfire and Sunda tsunami
    \item Fold 2: Moore tornado and Portugal wildfire
    \item Fold 3: Tuscaloosa tornado, Lower Puna volcano and Woolsey fire.
\end{itemize}
}
\vspace{-\parskip}
This choice is not random, but driven by the desire to balance test set sizes in terms of number of samples while including the three main damage-driving forces -- wind, fire and water -- in each split. %Dataset constituents, examples and statistics are also visualized in Fig.~\ref{fig:map}.

\begin{figure}
    \centering
    \includegraphics[width = 0.8\textwidth]{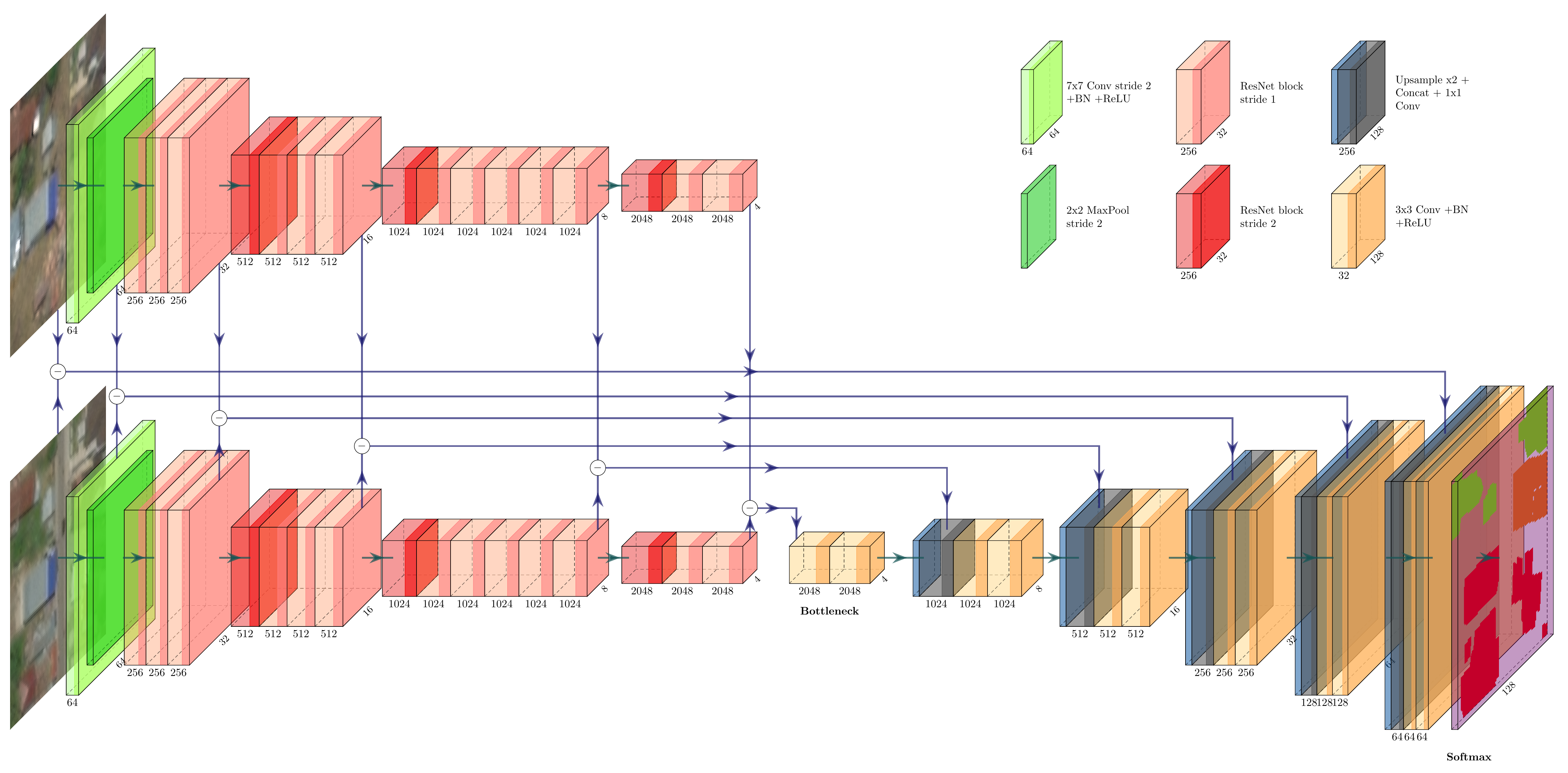}
    \vspace{-10pt}
    \caption{Our two-stream ResNet50 architecture.}
    \label{fig:tsrn50}
\end{figure}

% T-S RN50 + Train setup
\paragraph{Standard model: Two-stream ResNet50.}
We perform experiments with two model architectures. The first is a relatively standard approach slightly tweaked to match BDD. It takes the powerful ImageNet-pretrained ResNet50 backbone \citep{he2016} and uses it as the encoders in a two-stream U-Net \citep{ronneberger2015} architecture (Fig.~\ref{fig:tsrn50}). Here two-stream refers to having two seperate encoders, whose encodings are subtracted before feeding them into the decoder. The two encoders are pretrained on ImageNet, but their weights are not shared as we expect different image statistics pre- and post-disaster. We train the network over 60~epochs on the xBD training set cut into samples of size $256\times 256$\,px. We augment with random rotations, flips, rescaling, brightness, contrast, sharpness and color alterations. As the objective function we use a median-frequency balanced \citep{kampffmeyer2016} weighted cross-entropy loss. We optimize with ADAM \citep{kingma2014}, a learning rate of $10^{-4}$ and a batch size of~12.

% DualHRNet + Train setup
\paragraph{State-of-the-art model: Dual-HRNet.}
For comparison with the existing literature, we additionally use Dual-HRNet \citep{koo2020} for our experiments. We chose it because it is among the best-performing single-model entries in the xView 2 challenge (ranked 5\textsuperscript{th}) and a clean code base is available.\footnote{\url{https://github.com/DIUx-xView/xView2_fifth_place}} Dual-HRNet combines two ImageNet-pretrained HRNetV2 backbones \citep{wang2020} with intermediate fusion blocks and two heads, one for building localization and one for damage scoring. We follow the authors' training strategy, which trains the network over 500 epochs on lightly augmented $512 \times 512$\,px cutouts of samples in the xBD training set. It uses the Lovasz softmax loss \citep{berman2018}, which is optimized using stochastic gradient descent (SGD) and a polynomial learning rate policy. Since we have to use a smaller batch size of~6 (original paper:~32) due to limited GPU memory, we also decrease the initial learning rate from $0.05$ to~$0.01$.

% Domain adaptation: SWA, AdaBN
\paragraph{Methods to improve OOD generalization.} % Simple robustness-improving methods o.ä. Dann SWA als generalization method, AdaBN als Domain adapt --> dann in discussion schlussfolgerung Domain adapt könnte allgemein lohnenswert sein
%Domain adaptation is an area of research focussing on increasing model robustness to domain shifts by adapting the data or the model (or both) to overcome domain differences. 
We study two simple methods for improving OOD generalization: stochastic weight averaging \citep[SWA;][]{izmailov2018} and adaptive batch normalization~\citep[AdaBN;][]{li2018}.
The idea behind SWA is that loss surfaces tend to have multiple local minima in the area of the global minimum. This can be exploited by averaging the network parameters over multiple such local minima. In our implementation of SWA, we store the network parameters after each of the last 10 epochs (40 for Dual-HRNet) of training with constant learning rate of $0.005$ and then average them.

Batch normalization \citep[BN;][]{ioffe2015} layers are employed by many popular DNN architectures, including the ResNet50. They assume that the statistics from test samples are similar to those in training, so that the statistics from training can be used during testing. For OOD settings this may not be the case. AdaBN circumvents this problem and additionally might make the model robust to a potential shift in sample statistics from training to testing. AdaBN recalculates the statistics used in BN layers on the test set and uses these updated statistics during inference. AdaBN has recently proven useful for handling synthetic domain shifts \citep{schneider2020, nado2020}.

\begin{figure}[bt]
    \centering
    \includegraphics[width=\textwidth]{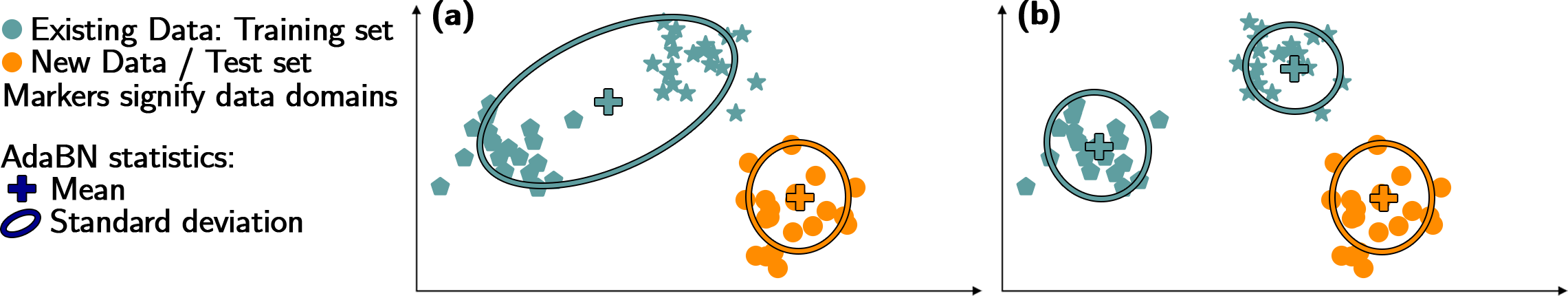}
    \caption{(a) Classic AdaBN in the multi-domain setting: Statistics during training are those of the mixture distribution, but at test-time single-domain statistics are used, rendering the test time setting atypical again. (b) Multi-domain AdaBN: During training, each batch contains just a single domain. As a result, single-domain statistics are used during training and training is representative of testing.}
    \label{fig:adabn}
\end{figure}

The classic AdaBN is defined for single-domain training. So model parameters are trained to work with the average BN statistics of the training set. In our case, the training set contains multiple domains. Therefore the BN statistics are not the statistics of individual domains but of the mixture distribution (see Fig.~\ref{fig:adabn}a). If at test time statistics of individual domains are used, these may not play well with the model's parameters. For instance, the variances during testing will be smaller (Fig.~\ref{fig:adabn}a) and activations will be scaled inappropriately. To circumvent this issue, we draw each minibatch from a single domain during training. As a result, the learned model parameters will be suited to work with single-domain BN statistics, which fits AdaBN at test time (Fig.~\ref{fig:adabn}b). We refer to this variation as \emph{multi-domain AdaBN}.

\section{Experiments}
\paragraph{Comparable performance to xView 2 challenge.}
We first assess performance of our two-stream ResNet50 on the standard IID test set of the xView 2 challenge (Table~\ref{tab:resultsallval}) and compare to a simple baseline and two state-of-the-art models. The Two-stream ResNet50 shows comparable performance to Dual-HRNet \citep{koo2020}, a state-of-the-art single-model approach. Note that performance can be improved by using heavy model ensembling (xView 2 winning solution by Victor Durnov), but we focus on single-model approaches due to limited compute resources. This result shows that our simple two-stream ResNet50 is a strong model in the standard IID setting.
{\renewcommand{\arraystretch}{1.2}
\setlength{\tabcolsep}{2.6pt}
\begin{table}[t]
    \caption{Model performance (F1 scores) on the xBD holdout set (IID). Bold denotes best performance among single-model approaches.}
    \small
    \label{tab:resultsallval}
    \centering
    \begin{tabularx}{\textwidth}{Xccccccccc}
         \toprule
         Model & Ensemble $\quad$ & xView2$\uparrow$ & $\quad$ & Loc & Dmg & 0 & 1 & 2 & 3 \\
         \midrule
         xView 2 baseline \citep{gupta2019} & \xmark & 0.27 & & 0.80 & 0.03 & 0.66 & 0.14 & 0.01 & 0.47 \\
         Two-stream ResNet50 (Ours) &  \xmark &0.77 & & 0.83 & \textbf{0.75} & \textbf{0.92} & 0.58 & \textbf{0.75} & \textbf{0.83} \\
         Dual-HRNet \citep{koo2020} & \xmark & \textbf{0.78} & & \textbf{0.87} & 0.74 & 0.90 & \textbf{0.59} & 0.74 & 0.81 \\
         \midrule
         xView 2 winning ensemble &  \checkmark & 0.81 & & 0.86 & 0.79 & 0.92 & 0.64 & 0.79 & 0.86 \\
        %  xView 2 baseline \citep{gupta2019} & 0.2654 & 0.8049 & 0.0342 & 0.6631 & 0.1435 & 0.0094 & 0.4657 \\
        %  Two-stream ResNet50 (Ours) & 0.7730 & 0.8331 & 0.7473 & 0.9215 & 0.5752 & 0.7534 & 0.8323 \\
        %  Dual-HRNet \citep{koo2020} & 0.7786 & 0.8659 & 0.7412 & 0.8985 & 0.5905 & 0.7378 & 0.8099 \\
        %  xView 2 winning ensemble & 0.8119 & 0.8635 & 0.7898 & 0.9234 & 0.6444 & 0.7859 & 0.8640 \\
         \bottomrule
    \end{tabularx}
    
\end{table}
}

% IID Gut -/-> OOD Gut
\paragraph{IID performance does not predict OOD performance.}
We now ask how these models fare on the more realistic OOD setting. To this end, we train both Two-stream ResNet50 and Dual-HRNet on the split used by \citet{gupta2020}. Consistent with their observation, we find a big generalization gap from disasters used during training to novel ones (Table~\ref{tab:resultsgupta}). Importantly, the IID performance is not a good indicator for OOD performance. While Two-stream ResNet50 and Dual-HRNet perform similarly well on IID data, the former generalizes poorly while the latter generalizes substantially better than the \citet{gupta2020} baseline, but still exhibits a generalization gap. We also observe a small drop in IID performance compared to full xBD training, likely due to a smaller training set.

{\renewcommand{\arraystretch}{1.2}
\setlength{\tabcolsep}{6pt}
\begin{table}[bt]
    \caption{Comparing performance with \citet{gupta2020}, table III.}
    \small
    \label{tab:resultsgupta}
    \centering
    \begin{tabularx}{\textwidth}{Xccccccccc}
            \toprule
        & \multicolumn{3}{c}{IID} & \multicolumn{5}{c}{OOD} & \\
         Model & xView2$\uparrow$ & Loc$\uparrow$ & Dmg$\uparrow$ &  & xView2$\uparrow$ & Loc$\uparrow$ & Dmg$\uparrow$ &  & Gap$\downarrow$\\
         \midrule
         \citet{gupta2020} & 0.66 & 0.79 & 0.60 & & 0.50 & 0.77 & 0.37 & & 0.16 \\
         Two-stream ResNet50 & 0.74 & 0.83 & 0.70 & & 0.44 & 0.77 & 0.30 & & 0.30\\%& 0.7382 & 0.8259 & 0.7007 & 0.4386 & 0.7722 & 0.2956 & -0.2996\\
         %\qquad +SWA & 0.7218 & 0.8371 & 0.6725 & 0.4954 & 0.7831 & 0.3720 & -0.2264\\
         %\qquad +AdaBN & 0.7166 & 0.8338 & 0.6664 & 0.5266 & 0.7731 & 0.4210 & -0.1900\\
         Dual-HRNet & 0.75 & 0.85 & 0.71 & & 0.61 & 0.80 & 0.53 & & 0.14\\%& 0.7508 & 0.8545 & 0.7064 & 0.6111 & 0.8005 & 0.5300 & -0.1397\\
         %\qquad +SWA & 0.7482 & 0.8535 & 0.7030 & 0.6157 & 0.8386 & 0.5201 & -0.1325\\
         \bottomrule
    \end{tabularx}
\end{table}
}

% Gupta split
% OOD-xBD
%\paragraph{OOD performance.}
%We compare the performance of Two-stream ResNet50 and DualHRNet with \citet{gupta2020}, the results are in table~\ref{tab:resultsgupta}. In line with them we find a big generalization gap on their challenging dataset. Unsurprisingly we also find a drop in model IID performance, likely due to fewer samples in the training set. 

\paragraph{The generalization gap is consistent across multiple test sets.}
To investigate whether the generalization gap we observe could be specific to the  challenging Gupta split (lack of training data) or due to a coincidental choice of a particularly hard OOD test set, we created OOD-xBD, (three additional, smaller OOD test sets; see Methods). With the additional training data, both models generalize better, but a significant gap remains (Table~\ref{tab:resultsdiff}). Also, in this setting the Two-stream ResNet50 outperforms Dual-HRNet on IID data while again scoring worse on OOD. These results show that the generalization gap observed above was not due to a coincidental split of the dataset and strengthens the point that IID performance is a poor indicator of OOD performance.

%The generalization gap also persists when switching to the easier yet more realistic and trustworthy OOD-xBD setting, results are in table~\ref{tab:resultsdiff}. DualHRNet remains better at generalization than Two-stream ResNet50. Possibly the differences in robustness comes from the different training regimes, more particularly it might be that the Two-stream ResNet50 has been trained too shortly.

{\renewcommand{\arraystretch}{1.2}
\setlength{\tabcolsep}{6pt}

\begin{table}[htb]
    
    \caption{Mean {\scriptsize($\pm$ standard deviation)} of xView2-Scores and their differences in the OOD-xBD setting.}
    \small
    \label{tab:resultsdiff}
    \centering
    \begin{tabularx}{\textwidth}{Xccc}
    \toprule
             Model & IID$\uparrow$ & OOD$\uparrow$ & Gap$\downarrow$ \\
             \midrule
             Two-stream ResNet50 & \textbf{0.74} \scriptsize{$\pm$0.01} & 0.60 \scriptsize{$\pm$0.01} & 0.13 \scriptsize{$\pm$0.02} \\
             %\qquad +SWA & 0.7472 \scriptsize{$\pm$0.0062} & 0.6209 \scriptsize{$\pm$0.0075} & -0.1263 \scriptsize{$\pm$0.0109} \\
             % \qquad +AdaBN & 0.6848 \scriptsize{$\pm$0.0208} & 0.5815 \scriptsize{$\pm$0.0336} & -0.1034 \scriptsize{$\pm$0.0543} \\
             DualHRNet & 0.73 \scriptsize{$\pm$0.01} & \textbf{0.67} \scriptsize{$\pm$0.02} & \textbf{0.05} \scriptsize{$\pm$0.02} \\
             
             %\qquad +SWA & 0.7142 \scriptsize{$\pm$0.0079} & 0.6584 \scriptsize{$\pm$0.0346} & -0.0558 \scriptsize{$\pm$0.0304} \\
             \bottomrule
    \end{tabularx}
\end{table}
}

\paragraph{Adaptive Batch Normalization and Stochastic Weight Averaging improve robustness.}
As our Two-stream ResNet50 generalized so poorly to the OOD setting, we wondered whether its OOD performance could be improved. We found that SWA and multi-domain AdaBN (see Methods) enhance its generalization capabilities (Table~\ref{tab:oodresults}). On the \citet{gupta2020} OOD test set, the Two-stream ResNet50 improved substantially by combining multi-domain AdaBN and SWA, with the model now scoring at $0.59$. The same is true for Dual-HRNet, which is also improved substantially on the Gupta split. Here multi-domain AdaBN seems to be more important than SWA, presumably because of the different training regimes used for Two-stream ResNet50 and Dual-HRNet.

The results are confirmed in the OOD-xBD setting. Here, as more training data is available, gains are smaller. Still, the Two-stream ResNet50 OOD performance is improved by $0.05$ using the combination of methods. For the Dual-HRNet, the performance gain given by SWA vanishes and here just multi-domain AdaBN improves OOD generalization mildly. The Dual-HRNet with multi-domain AdaBN in the OOD-xBD setting is the network with the overall lowest generalization gap of $0.04$. 
%Applying SWA improves both IID and OOD scores, presumably because it shifts model weights to more robust minima. 

%The classic AdaBN~\citep{schneider2020} that simply adapts the BN statistics to each target domain actually decreases IID performance and only marginally improves OOD scores. We believe this happens for the following reason: under the classic regime, model parameters are trained to work with the average BN statistics of the training set, which includes multiple domains. Therefore, the BN statistics are not the statistics of individual domains but of the mixture distribution (see Fig.~\ref{fig:iidood}b). If at test time statistics of individual domains are used, these do not play well with the model's parameters. For instance, the variances during testing will be smaller (Fig.~\ref{fig:iidood}b) and activations will be scaled inappropriately. To circumvent this issue, we change the training method such that each batch only contains samples from a single domain. Then model parameters will be suited to work with single-domain BN statistics, which fits AdaBN at test time. We refer to this variation as \emph{multi-domain AdaBN}. It recovers some of the IID performance and, importantly, boosts OOD performance substantially. Additionally using SWA improves OOD generalization even further, so that Two-stream ResNet50 is now competitive with Dual-HRNet. The same methods do also improve the performance of DualHRNet, still a generalization gap remains.

{\renewcommand{\arraystretch}{1.2}
\setlength{\tabcolsep}{6pt}
\begin{table}[ht]
    \caption{Using multi-domain AdaBN and SWA significantly improve OOD generalization. Score: xView 2 Score on the respective OOD test set, for OOD-xBD it is the mean {\scriptsize($\pm$ standard deviation)} of the three folds. Gap: generalization gap of a model calculated as the difference between that models' mean IID and OOD scores. Gain: improvement in OOD score relative to same model's baseline. Bold: best score per model class and dataset. Underlined: best overall score per dataset.}
    \small
    \label{tab:oodresults}
    \centering
    \begin{tabularx}{\textwidth}{Xcccccccc}
            \toprule
        & \multicolumn{5}{c}{\citet{gupta2020} OOD} & \multicolumn{3}{c}{OOD-xBD} \\
         Model & $\quad$ & Score$\uparrow$ & Gap$\downarrow$ & Gain$\uparrow$ & $\quad$ & Score$\uparrow$ & Gap$\downarrow$ & Gain$\uparrow$  \\
         \midrule
         Two-stream ResNet50 & & 0.44 & 0.30 & -- & & 0.60 \scriptsize{$\pm$0.01} & 0.14 & -- \\%0.13 \scriptsize{$\pm$0.02} & 0.00\\
         \quad +SWA & & 0.50 & 0.23 & 0.06 & & 0.62 \scriptsize{$\pm$0.01} & 0.13 & 0.02 \\%0.13 \scriptsize{$\pm$0.01} & 0.02 \scriptsize{$\pm$0.01}\\
         %\quad +classic AdaBN &  & 0.53 & 0.19 & 0.09 & & 0.58 \scriptsize{$\pm$0.03} & 0.10 & -0.02 \\%0.10 \scriptsize{$\pm$0.05} & -0.02 \scriptsize{$\pm$0.03} \\
         %\quad +classic AdaBN +SWA & & 0.51 & 0.22 & 0.07 & & 0.59 \scriptsize{$\pm$0.04} & 0.12 & -0.01 \\%0.13 \scriptsize{$\pm$0.05} & -0.02 \scriptsize{$\pm$0.04}\\
         \quad +multi-domain AdaBN & & 0.52 & 0.17 & 0.07 & & 0.62 \scriptsize{$\pm$0.05} & 0.08 & 0.02 \\%0.12 \scriptsize{$\pm$0.08} & -0.04 \scriptsize{$\pm$0.12}\\
         \quad +multi-domain AdaBN +SWA & & \textbf{0.59} & 0.15 & 0.15 & & \textbf{0.65} \scriptsize{$\pm$0.03} & 0.07 & 0.05 \\%0.05 \scriptsize{$\pm$0.02} & 0.08 \scriptsize{$\pm$0.02}\\
         Dual-HRNet & & 0.61 & 0.14 & -- & & 0.67 \scriptsize{$\pm$0.02} & 0.06 & -- \\%0.05 \scriptsize{$\pm$0.02} & 0.00\\ 
         \quad +SWA & & 0.62 & 0.13 & 0.00 & & 0.66 \scriptsize{$\pm$0.03} & 0.05 & -0.01 \\%0.06 \scriptsize{$\pm$0.03} & -0.02 \scriptsize{$\pm$0.01}\\
         %\quad +classic AdaBN & & 0.64 & 0.07 & 0.03 & & 0.66 \scriptsize{$\pm$0.05} & 0.05 & -0.01 \\%0.06 \scriptsize{$\pm$0.04} & -0.02 \scriptsize{$\pm$0.03}\\
         %\quad +classic AdaBN +SWA & & 0.63 & 0.08 & 0.02 & & 0.67 \scriptsize{$\pm$0.04} & 0.05 & 0.01 \\%0.04 \scriptsize{$\pm$0.04} & 0.00 \scriptsize{$\pm$0.03}\\
         \quad +multi-domain AdaBN & & 0.67 & 0.04 & 0.06 & & \underline{\textbf{0.69}} \scriptsize{$\pm$0.03} & 0.04 & 0.02 \\%0.04 \scriptsize{$\pm$0.02} & 0.02 \scriptsize{$\pm$0.02}\\
         \quad +multi-domain AdaBN +SWA & & \underline{\textbf{0.68}} & 0.05 & 0.07 & & 0.68 \scriptsize{$\pm$0.03} & 0.06 & 0.01 \\%0.06 \scriptsize{$\pm$0.03} & 0.01 \scriptsize{$\pm$0.02}\\
         \bottomrule
    \end{tabularx}
\end{table}
}

\section{Discussion}
Building damage detection is a natural multi-domain setting that should be evaluated on OOD test sets. We show that IID performance does not predict OOD performance in BDD. Following \citet{geirhos2020}, we recommend that OOD testing should become the rule rather than the exception also in building damage detection. The OOD-xBD test setting we designed tests exactly this property. When testing OOD, we find evidence that existing models are subject to a significant generalization gap. One possible way of tackling this gap is using known robustness methods. We studied two methods, multi-domain AdaBN, a modified version of the simple domain adaptation technique AdaBN, as well as SWA. Both are helpful in increasing model robustness to domain shifts. Thus, building damage detection could become an interesting real-world test case for evaluating and improving domain adaptation and domain generalization methods.

{%\small
\paragraph{Author contributions.}
VB: conducted experiments, designed figures and wrote manuscript. AE: revised manuscript, wrote final version, steered research, supervised and provided resources.\\
\textbf{Acknowledgments.}
We are thankful for invaluable help, comments and discussions to all lab members, especially to Max Burg, Santiago Cadena, Kai-Hendrik Cohrs, Timo Lüddecke, Claudio Michaelis, Marita Schwahn and Marissa Weis. We thank Max Wardetzky for co-supervising at an earlier stage of this work. We estimate this project has caused between $1$ and $3$ tons of carbon emissions, which we commit to offset.
}
%\newpage
{%\small
\bibliography{refsurl}
\bibliographystyle{iclr2021_conference}
}
\end{document}